\documentclass{article}

\usepackage{arxiv}

\usepackage[utf8]{inputenc}
\usepackage[T1]{fontenc}
\usepackage{hyperref}   
\usepackage{url}        
\usepackage{booktabs}   
\usepackage{amsfonts}   
\usepackage{nicefrac}   
\usepackage{microtype}  
\usepackage{cleveref}   
\usepackage{lipsum}     
\usepackage{graphicx}
\usepackage{natbib}
\usepackage{doi}
\usepackage{float}

\title{Unveiling Document Structures with YOLOv5 Layout Detection}

\author{ 
    \href{https://orcid.org/0000-0003-4610-0456}{\includegraphics[scale=0.06]{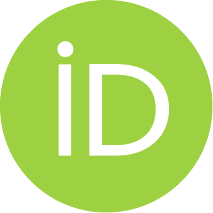}\hspace{1mm} Herman Sugiharto} \\
	Department of Informatics\\
	Siliwangi University\\
	Tasikmalaya, Indonesia \\
	\texttt{177006045@student.unsil.ac.id} \\
	\and
	Yorisa Silviana \\
	Department of Informatics\\
	Siliwangi University\\
	Tasikmalaya, Indonesia \\
	\texttt{2170060502@student.ac.id} \\
	\AND
	Yani Siti Nurpazrin \\
	Department of Informatics\\
	Siliwangi University\\
	Tasikmalaya, Indonesia \\
	\texttt{21006047@unsil.ac.id} \\
}

\hypersetup{
pdftitle={Unveiling Document Structures with YOLOv5 Layout Detection},
pdfsubject={q-cs.AI, q-cs.AI},
pdfauthor={Herman Sugiharto, Yorissa Silviana, Yani Siti Nurpazrin},
pdfkeywords={layout detection, unstructured data, YOLOv5},
}

\begin{document}
\maketitle

\begin{abstract}
	The current digital environment is characterized by the widespread presence of data, particularly unstructured data, which poses many issues in sectors including finance, healthcare, and education. Conventional techniques for data extraction encounter difficulties in dealing with the inherent variety and complexity of unstructured data, hence requiring the adoption of more efficient methodologies. This research investigates the utilization of YOLOv5, a cutting-edge computer vision model, for the purpose of rapidly identifying document layouts and extracting unstructured data.

	The present study establishes a conceptual framework for delineating the notion of "objects" as they pertain to documents, incorporating various elements such as paragraphs, tables, photos, and other constituent parts. The main objective is to create an autonomous system that can effectively recognize document layouts and extract unstructured data, hence improving the effectiveness of data extraction.

	In the conducted examination, the YOLOv5 model exhibits notable effectiveness in the task of document layout identification, attaining a high accuracy rate along with a precision value of 0.91, a recall value of 0.971, an F1-score of 0.939, and an area under the receiver operating characteristic curve (AUC-ROC) of 0.975. The remarkable performance of this system optimizes the process of extracting textual and tabular data from document images. Its prospective applications are not limited to document analysis but can encompass unstructured data from diverse sources, such as audio data.

	This study lays the foundation for future investigations into the wider applicability of YOLOv5 in managing various types of unstructured data, offering potential for novel applications across multiple domains.

\end{abstract}


\section{Introduction}

In the contemporary and dynamic digital age, there has been a substantial rise in the generation and utilization of data. Unstructured data, which refers to data that does not possess a predetermined 
format, holds significant importance inside diverse domains including banking, healthcare, and education.\cite{Adnan2019a}.
A significant portion of the data contained in documents is found in unstructured formats and exhibits variability in terms of its style and presentation, hence posing difficulties in the extraction of crucial information.\cite{Adnan2019b}. 
When faced with these variances and complexities, conventional methods of data extraction frequently demonstrate ineffectiveness and inefficiency \cite{Zaman2020}.
In order to tackle this matter, the utilization of technologies such as artificial intelligence and computer vision has facilitated the process of data extraction and processing. 
Nevertheless, there exists potential for enhancement in terms of velocity, precision, and effectiveness. \cite{Diwan2022}.

Detecting objects is a fundamental task in computer vision with numerous applications, including layout detection. 
Throughout the years, the YOLO (You Only Look Once) line of models has emerged as a prominent solution for real-time object identification, 
renowned for their exceptional speed and accuracy \cite{Jimenez2022Multi}. YOLOv5, the most recent edition of the YOLO family, demonstrates notable advancements 
in accuracy and precision when compared to its previous versions. While YOLOv4 shown remarkable performance, YOLOv5 has been rigorously crafted to augment accuracy while maintaining efficient inference speed \cite{Kaur2022Tools}\cite{Arifando2023Improved}. 
Through a combination of architectural refinements, novel data augmentation techniques, and a carefully curated training process, 
YOLOv5 accomplishes superior object detection capabilities \cite{Hussain2023YOLO}. 

This study's primary objective is to investigate and enhance the application of techniques for identifying document layouts and extracting unstructured data 
using the YOLOv5 framework. This study defines "objects" as the many components found within documents, including but not limited to paragraphs, tables, photographs, and other similar items. 
The primary aim of this study is to develop and deploy a system capable of autonomously identifying document layouts and efficiently and precisely extracting unstructured data from these documents. 
This study is expected to provide a valuable contribution towards enhancing the efficacy of unstructured data extraction.

\section{Related Work}
Numerous studies on layout detection and the application of the YOLOv5 architecture have been utilized in the past. 
In a meticulously executed research project conducted by \cite{Pfitzmann2022}, the academic community was introduced to the revolutionary DocLayNet dataset. The dataset presented below signifies a significant transformation in the domain of document layout research, providing an extensive collection of meticulously annotated document layouts. 
It consists of an astounding total of 1,107,470 meticulously annotated objects, encompassing a wide range of diverse object classes, including but not limited to text, images, mathematical formulas, code snippets, page headers and footers, and intricate tabular structures.
In contrast, the research undertaken by \cite{Pillai2021} followed a different research path, focusing on data derived from the complex field of the oil and gas business. 
The study utilized advanced transformer topologies to address the challenging problem of detecting and extracting layout components that are embedded within intricate papers from this particular domain.

The YOLOv5 framework has been employed in a multitude of computer vision research endeavors, encompassing several domains such as 
object recognition \cite{Diwan2022, Yue2022, Kitakaze2020}, object tracking \cite{Alvar2018, ProfDrKhalilIAlsaif2021,Kumari2021}, 
and video analysis \cite{Wang2022,Gu2022}.
In the aforementioned experiments, YOLOv5 has exhibited a notable level of precision in conjunction with its user-friendly nature.

In this exhaustive study, the research team has developed a sophisticated system that goes beyond layout detection; it incorporates the intricate task of layout extraction guided by meticulously predefined classes. 
At the core of this robust system lies YOLOv5, an advanced deep learning framework that serves as the layout detector. 
Its presence and performance in the system contribute significantly to the overarching framework's exceptional precision and efficacy.

The primary objective of this research is to revolutionize the processing of unstructured data, with a particular concentration on PDF documents generated from scanned sources. The documents in question provide a significant obstacle for traditional methods of extracting text from PDF files, since they are typically hindered by the complexities of scanned images. 
The unique approach employed by the study team holds the potential to surpass the existing constraints, providing a powerful solution to the challenging endeavor of efficiently extracting information from these texts. As we progress further into the era of digital transformation, the advances made by this research hold the promise of substantial advances in document processing, bridging the divide between unstructured data and actionable insights.

\section{Methodology}
The research is a quantitative study with an experimental approach. The experimental approach is chosen because the aim of this research is to determine 
the cause-and-effect relationships among existing variables such as datasets, model architectures, and model parameters (Williams, 2007).

The novelty targeted by this proposed research lies in the utilization of YOLOv5 for detecting layouts within a document.

\paragraph{Literature Review}
The literature survey was undertaken in order to gain a comprehensive understanding of the concepts and theories that are relevant to the research. This includes exploring the theoretical foundations of the YOLO architecture, examining the process of data labeling, and investigating the techniques used for layout detection. The data was obtained from secondary sources, including online platforms, academic publications, electronic books, scholarly papers, and other relevant materials. 
Furthermore, in the literature review phase, a comprehensive examination of prior scholarly articles was conducted to assess the research that pertains to the present research subject.

\paragraph{Problem Definition}
Through an examination of prior research, several gaps or weaknesses within these studies were uncovered, hence highlighting opportunities for prospective enhancements. 
After identifying gaps or weaknesses, the researchers generated research questions to establish the aims of the next study.

\paragraph{Data Collection}
During this phase, the data underwent preparation in order to train the forthcoming layout detection model. The dataset included of photos depicting the layout of documents sourced from a variety of academic journals. 
The data was subsequently annotated using Label Studio, employing pre-established categories.

\paragraph{Model Training}
During this stage, the existing YOLOv5 architecture was trained using optimal parameters to produce an appropriate model. 
The model was trained using the provided hardware and labeled data.

\paragraph{Model Evaluation}
During this phase, the trained model was subjected to several tests utilizing the pre-existing provided data. 
The evaluation process additionally incorporated manual human assessment in order to augment the validity of the evaluation data. 
The evaluation process involved the utilization of metrics such as accuracy, precision, and F1 score for the purpose of calculations.

\paragraph{Conclusion}
Drawing conclusions provided an overview of the data analysis and model evaluation, 
encompassing the entirety of the research.

\section{Results and Discussion}
\subsection{Base Model}
YOLO was initially proposed by \cite{redmon2016look} in 2016. 
This method gained recognition for its real-time processing speed of 45 frames per second. 
Simultaneously, the method maintained competitive performance and even achieved state-of-the-art results on popular datasets.

YOLOv5 is designed for fast and accurate real-time object detection. This algorithm offers several performance enhancements 
compared to its previous versions \cite{redmon2016yolo9000,redmon2016look,redmon2018yolov3}, including improved speed and detection capabilities. 
One of the key advantages of YOLOv5 is its ability to conduct object detection swiftly on resource-constrained devices such as CPUs or mobile devices. 
This enables researchers or academics to perform real-time object detection rapidly without sacrificing accuracy \cite{Jocher2022Zenodo}.

\begin{figure}[H]
	\centering
	\includegraphics[width=0.5\textwidth]{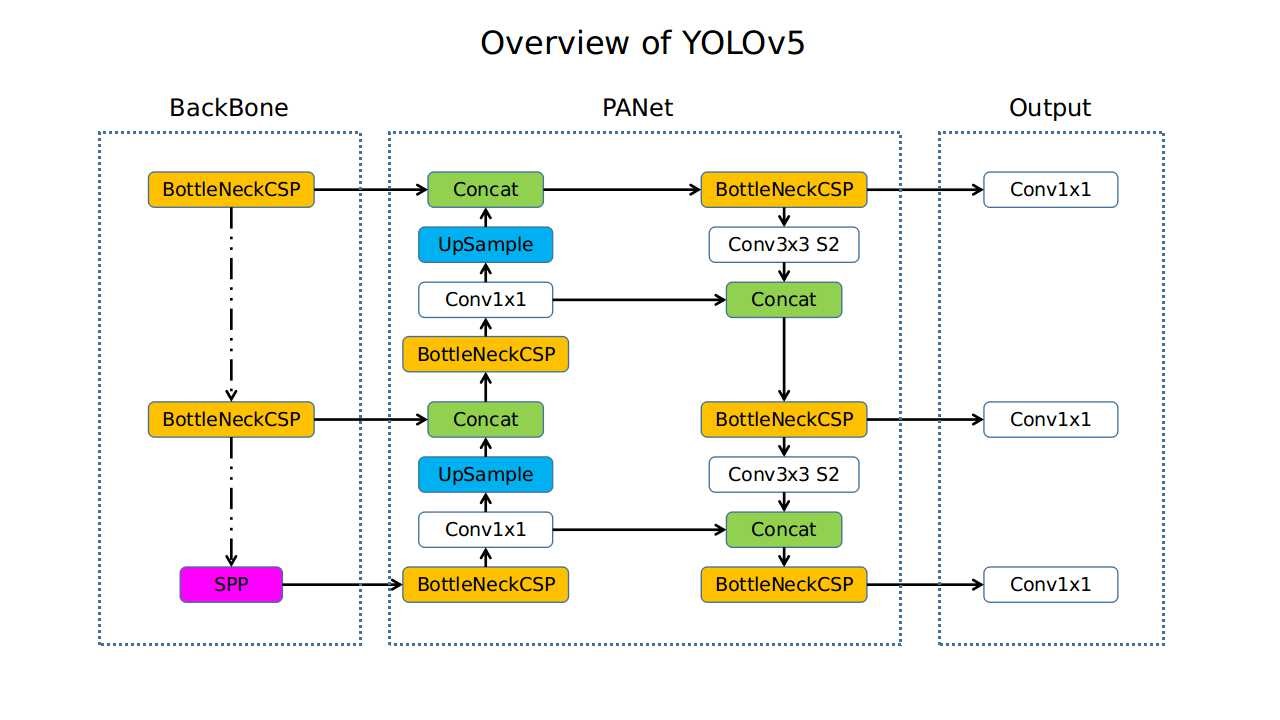}
	\caption{YOLOv5 architecture \cite{Jocher2022Zenodo}.}
	\label{fig:fig2}
\end{figure}

The architectural design of YOLOv5, as illustrated in Figure \ref{fig:fig2}, showcases its segmentation into three main components: Backbone, PANet, and Output.
The Backbone, alternatively referred to as the feature extractor, is a crucial component within a network that is tasked with extracting fundamental elements from the input image. 
The YOLOv5 model incorporates the CSPDarknet53 architecture as its underlying framework. 
The Path Aggregation Network (PANet) is a key element of the YOLOv5 framework, designed to effectively aggregate information from many scales. 
The PANet architecture facilitates the integration of contextual information from many scales, hence enhancing the ability to recognize objects of varying sizes. 
The YOLOv5 model produces a result of several bounding boxes and corresponding class labels, representing the detected objects in the given image. 
According to Jin (2022), bounding boxes are utilized to establish the precise coordinates and dimensions of objects within an image, 
while class labels serve to identify the specific category to which the identified object belongs.

\subsection{Layout Detection}
\begin{figure}[H]
	\centering
	\includegraphics[width=0.5\textwidth]{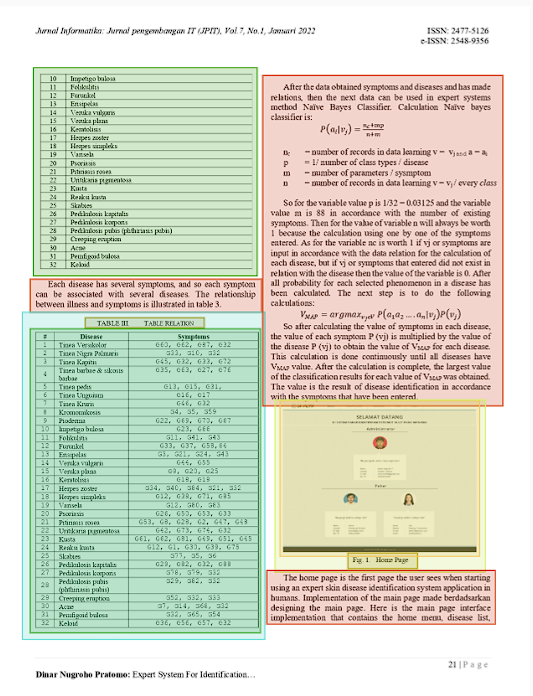}
	\caption{Document Layout.}
	\label{fig:fig3}
\end{figure}

\begin{figure}[H]
	\centering
	\includegraphics[width=0.5\textwidth]{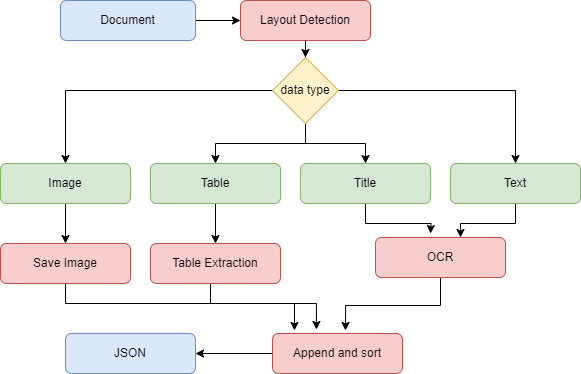}
	\caption{Layout Detection Flow.}
	\label{fig:fig4}
\end{figure}

The technique of \emph{Layout Detection} is utilized to ascertain the configuration of elements within a document \cite{Vitagliano2022}. 
In this study, the term "layout" refers to the various components that comprise the structure of a layout, including titles, text, photos, captions, and tables, as seen in Figure \ref{fig:fig3}.
The data extraction process for detected documents is determined based on the specific type of data contained inside them. The process of extracting data is depicted in Figure \ref{fig:fig4}.

The extraction components used in this research are as follows:

\paragraph{Optical Character Recognition (OCR)} This method is employed to transform text data present in 
scanned documents into editable and searchable text \cite{Billah2015}. 
The OCR framework used in this research is Tesseract. 
Tesseract is a framework developed by Google for optical character recognition needs, offering ease of use \cite{Smith2007}.

\paragraph{Table extraction} encompasses two components, table structure recognition and OCR. 
Table structure recognition is used to detect the structure of tables, including rows, columns, and cells. 
The PubTables-1M model \cite{smock2021pubtables1m} is utilized for this purpose. 
This model accurately analyzes tables originating from images.

The extracted data will be combined into a JSON format and sorted based on the coordinate positions of the data components. 
Consequently, the obtained data will include component coordinates (x1, y1, x2, y2), component classes (such as text, tables, etc.), 
and data, as depicted in Figure \ref{fig:fig4}.

\subsection{Dataset}
The dataset included in this study comprises 153 PDF pages that have been transformed from diverse sources, such as books and sample journals. 
The data was subsequently tagged utilizing Label Studio \cite{LabelStudio} with the subsequent classes:

\begin{table}[H]
    \caption{Data Classes.}
    \centering
    \begin{tabular}{ll}
        \toprule
        Class & Description \\
        \midrule
		Title & Attribute referring to the book title \\
		Text & Attribute referring to the text within the book \\
		Image & Attribute indicating images on the book page \\
		Caption & Attribute for captions of images or tables \\
		Image\_caption & Group box for images and captions \\
		Table & Attribute for tables in the book \\
		Table\_caption & Group box for tables and captions \\
        \bottomrule
    \end{tabular}
    \label{tab:table1}
\end{table}

\begin{figure}[H]
	\centering
	\includegraphics[width=0.4\textwidth]{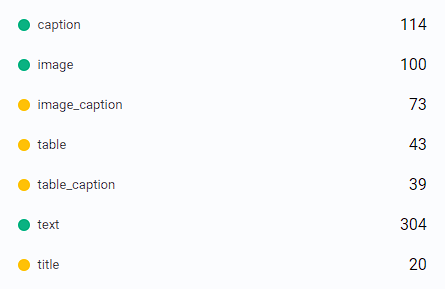}
	\caption{Data train class.}
	\label{fig:fig6}
\end{figure}

Each page within the used dataset has a varying number of classes due to the distinct structures of each page. 
The classes for the training data are indicated as shown in Figure \ref{fig:fig6}.

The training data consists of 143 layout image data, while the test data comprises 10 layout image data, 
with data classes visible in Figure 8.

\begin{figure}[H]
	\centering
	\includegraphics[width=0.5\textwidth]{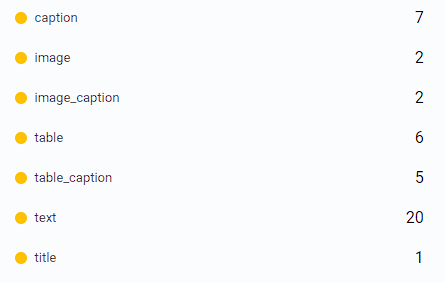}
	\caption{Data test class.}
	\label{fig:fig7}
\end{figure}

\subsection{Training Model}

When conducting training, the parameters employed are outlined in Table \ref{tab:table2}.

\begin{table}[H]
    \caption{Data Classes}
    \centering
    \begin{tabular}{ll}
        \toprule
		Parameter & Value \\
        \midrule
		Model variant & YOLOv5 S \\
		Epoch & 500 \\
		Image Size & 640 \\
		Patience & 100 \\
		Cache & RAM \\
		Device & GPU \\
		Batch size & 32 \\
        \bottomrule
    \end{tabular}
    \label{tab:table2}
\end{table}

The environment utilized to execute the training is Google Colab Pro, with specifications as provided in Table \ref{tab:table3}.

\begin{table}[H]
    \caption{Hardware specifications}
    \centering
    \begin{tabular}{ll}
        \toprule
		Hardware & Specification \\
        \midrule
		CPU	& 2 x Intel Xeon CPU @ 2.20GHz \\
		GPU	& Tesla P100 16 GB \\
		RAM	& 27 GB \\
		Storage &	129 GB available \\
        \bottomrule
    \end{tabular}
    \label{tab:table3}
\end{table}

\subsection{Evaluation Metric}
Evaluation metrics are tools used to measure the quality and performance of machine learning models \cite{Thambawita2020}. 
Some of the metrics used include mAP50, mAP50-95, Precision, Recall, Box Loss, Class Loss, and Object Loss.

\paragraph{Precision} is the ratio of true positive predictions (TP) to the total number of positive predictions $(TP + FP)$. 
Precision is used to measure the quality of positive predictions by the model \cite{Heyburn2018}. 
Precision is defined as shown in Equation (1):

\begin{equation}
P = \frac{TP}{TP + FP}
\end{equation}

\paragraph{Recall} is the ratio of true positive predictions (TP) to the total number of actual positives $(TP + FN)$. 
Recall is used to measure the model's ability to find all positive samples \cite{Wang2022}. 
Recall is defined as shown in Equation (2):

\begin{equation}
R = \frac{TP}{TP + FN}
\end{equation}

\paragraph{mAP50} The average of the Average Precision (AP) is calculated by considering all classes. 
A detection is deemed correct if the Intersection over Union (IoU) between the predicted bounding box and the ground truth is 0.5 or higher. 
The aforementioned metric offers an assessment of the model's effectiveness in object detection, allowing for a certain degree of flexibility 
in terms of mistakes related to object placement and bounding box dimensions \cite{Heyburn2018}.

\paragraph{mAP50-95} The assessment metric employed in object detection tasks is frequently utilized inside competitive settings, such as the COCO (Common Objects in Context) challenge. 
The metric being referred to is the mean Average Precision (mAP) calculated across different Intersection over Union (IoU) criteria. These thresholds range from 0.5 to 0.95, with an increment of 0.05 \cite{Thambawita2020}.

\paragraph{Box Loss} The metric referred to as box loss, or alternatively localization loss, evaluates the accuracy of a model's predictions regarding object bounding boxes. 
The calculation often involves determining the disparity between the predicted bounding box coordinates generated by the model and the corresponding actual (ground truth) bounding box coordinates. 
Two often employed metrics in this context are Mean Squared Error (MSE) and Intersection over Union (IoU). \cite{Wang2022}.

\paragraph{Class Loss} The metric of class loss evaluates the model's ability to accurately forecast object classes. 
The calculation typically involves determining the discrepancy between the anticipated probability of class membership as estimated by the model and the true classes as determined by the ground truth. 
Cross-Entropy Loss is a frequently employed metric in this context \cite{Wang2022}.

\paragraph{Object Loss} The metric of object loss evaluates the model's ability to accurately forecast the existence of objects. 
In models like as YOLO, the prediction of the presence or absence of an object at the center of each cell in the visual grid is made. 
The calculation of object loss involves determining the discrepancy between the anticipated probability of object presence as determined by the model and the actual presence of the object, as indicated by the ground truth \cite{Heyburn2018}.

\subsection{Training Results}
\begin{figure}[H]
	\centering
	\includegraphics[width=0.5\textwidth]{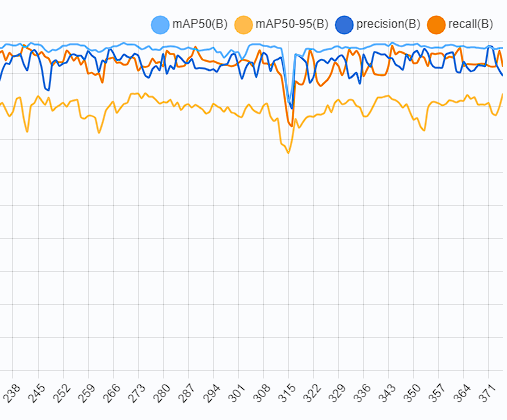}
	\caption{Training Model Metric Graph}
	\label{fig:metric}
\end{figure}

The training results yield metric values as shown in Table \ref{tab:metric}, 
indicating mAP50, mAP50-95, Precision, and Recall scores. 
Figure \ref{fig:metric} illustrates the metric graph for iterations 238 to 381.

\begin{table}[H]
    \caption{Training Model Metric}
    \centering
    \begin{tabular}{ll}
        \toprule
		Metric & Value \\
        \midrule
		mAP50 & 0.97 \\
		mAP50-95 & 0.801 \\
		Precision & 0.911 \\
		Recall & 0.971 \\
        \bottomrule
    \end{tabular}
    \label{tab:metric}
\end{table}

These results show that the model training has achieved a sufficiently high accuracy for predicting the provided document layouts. The results also indicate that the training data stopped at epoch 381 due to achieving satisfying accuracy and no further improvement, leading to early stopping of the model.

\begin{figure}[H]
\centering
\includegraphics[width=0.5\textwidth]{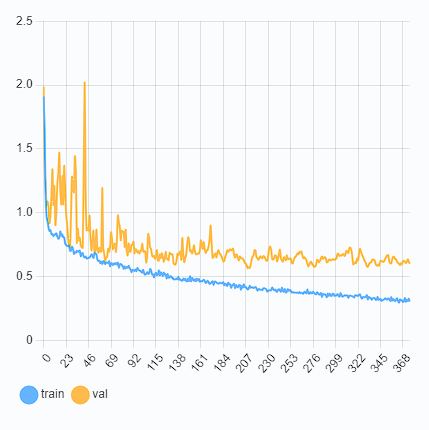}
\caption{Box Loss Metric Results}
\label{fig:box_loss}
\end{figure}

Box Loss as depicted in Figure \ref{fig:box_loss} has values of 0.308 during the training process and 0.636 during validation. These results indicate that the model can predict object bounding boxes well with low data loss.

\begin{figure}[H]
\centering
\includegraphics[width=0.5\textwidth]{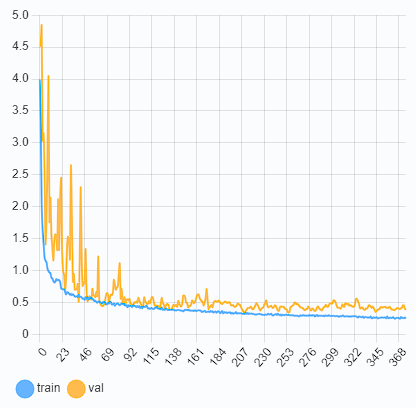}
\caption{Class Loss Metric Results}
\label{fig:class_loss}
\end{figure}

The model training yields small class loss values of 0.245 during training and 0.383 during validation, as shown in Figure \ref{fig:class_loss}. This demonstrates the model's ability to predict classes from the given layouts.

\begin{figure}[H]
\centering
\includegraphics[width=0.5\textwidth]{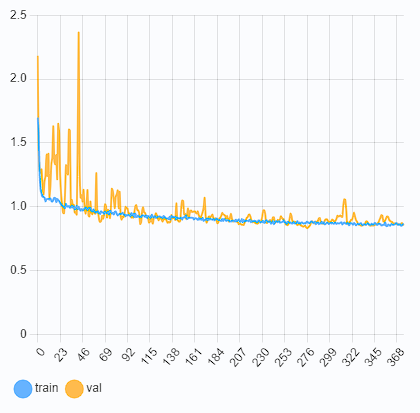}
\caption{Object Loss Metric Results}
\label{fig:object_loss}
\end{figure}

The Object Loss metric refers to the model's ability to detect objects before predicting their classes and bounding boxes. The training value is 0.863, and the validation value is 0.85, as shown in Figure \ref{fig:object_loss}.

\begin{figure}[H]
\centering
\includegraphics[width=0.5\textwidth]{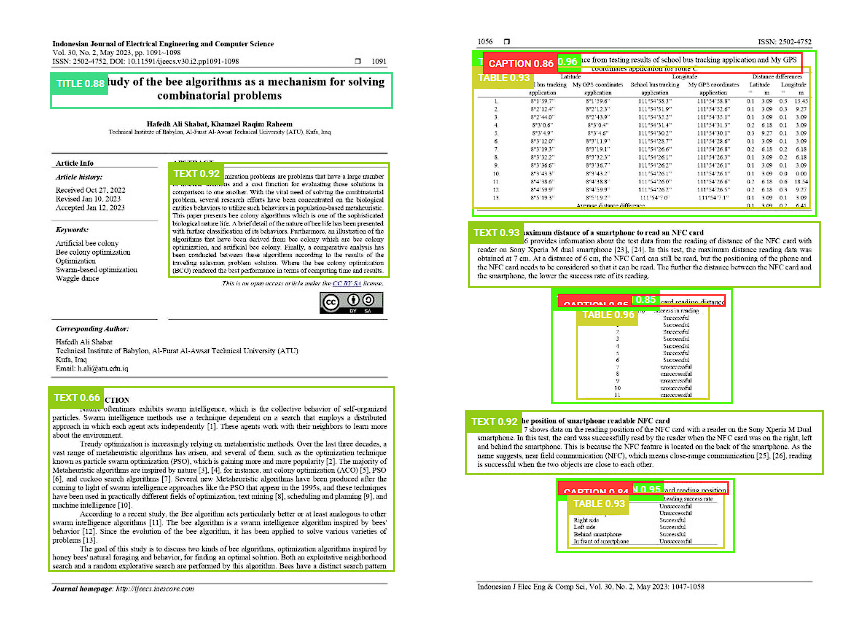}
\caption{Object Detection Results}
\label{fig:detection}
\end{figure}

The results of the extraction process are exemplified in Figure \ref{fig:detection}, demonstrating accurate predictions with high speed.

\begin{figure}[H]
\centering
\includegraphics[width=0.5\textwidth]{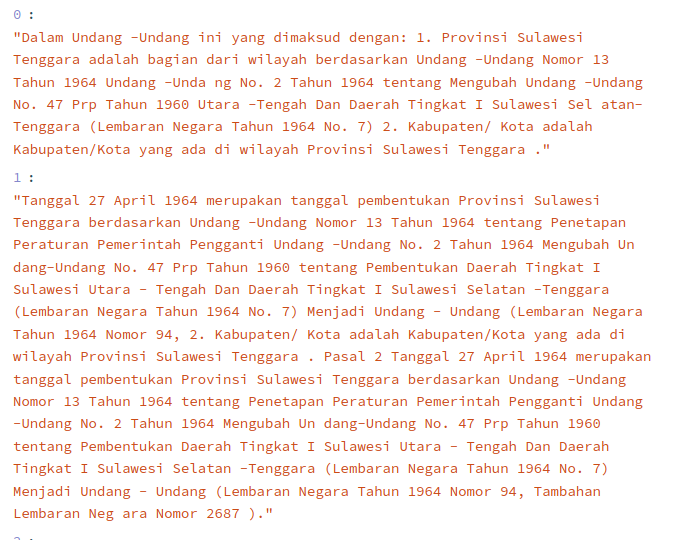}
\caption{Text Extraction Results}
\label{fig:extraction}
\end{figure}

Extraction results using regulation page data are shown in Figure \ref{fig:extraction}, aligning with the original data. The average extraction speed is 0.512 per page.

The outcomes of the detection and extraction process provide evidence that the model successfully meets the criteria for functioning as an unstructured document detector and extractor.

\section{Conclusions}
The utilization of YOLOv5 in document layout identification tasks has demonstrated significant efficacy, resulting in a notable accuracy rate accompanied with precision values of 0.91 and recall values of 0.971. 
The exceptional performance of this model has facilitated its ability to identify and retrieve textual and tabular data from document images, hence accelerating the typically arduous task of extracting data from scanned documents. 
The capabilities of YOLOv5 can be further expanded beyond the analysis of document layout, presenting opportunities for exciting future study. 
This entails exploring the possibilities of utilizing many forms of unstructured data, encompassing not just documents and photographs but also audio data analysis. This avenue has significant opportunities for a broad spectrum of applications.

\bibliographystyle{unsrtnat}
\bibliography{references}

\end{document}